\documentclass{article}

\usepackage[final]{neurips_2019}
\usepackage[utf8]{inputenc}
\usepackage[T1]{fontenc}
\usepackage{hyperref}
\usepackage{url}
\usepackage{booktabs}
\usepackage{amsfonts}
\usepackage{nicefrac}
\usepackage{microtype}
\usepackage[hang,flushmargin]{footmisc}

\usepackage{amsmath}
\usepackage{amssymb}
\usepackage{amsthm}
\usepackage{graphicx}
\theoremstyle{definition}
\newtheorem{definition}{Definition}[section]

\newcommand{\ra}{\rightarrow}
\newcommand{\E}[2][ ]{\mathbb{E}_{#1} \left [ #2\right ]}

\title{Defining and Evaluating Fair Natural Language Generation}

\author{
Catherine Yeo \\
Harvard University\\
\texttt{cyeo@college.harvard.edu} \\
\And
Alyssa Chen \\
Harvard University\\
\texttt{alyssachen@college.harvard.edu} \\
}

\begin{document}

\maketitle
\begin{abstract}
  Our work focuses on the biases that emerge in the natural language generation (NLG) task of sentence completion. In this paper, we introduce a framework of fairness for NLG followed by an evaluation of gender biases in two state-of-the-art language models. Our analysis provides a theoretical formulation for biases in NLG and empirical evidence that existing language generation models embed gender bias.  
\end{abstract}

\section{Introduction}
Quantifying fairness is a major consideration in the ethical use of natural language processing (NLP) tasks such as machine translation and word embedding. Standard machine learning models succeed through training on corpora of text usually written by humans. However, this training process results in human-like semantic biases in language models \cite{caliskan}. Language models can not only embed bias, but have been shown in some NLP tasks to amplify existing bias \cite{zhao}, a result which further underscores the importance of detecting biases and requiring fairness. 

State-of-the-art natural language generation models exhibit such human-like biases. \newcite{sheng} found that when given the prompts of ``The man worked'' and ``The woman worked'', OpenAI's GPT-2 \cite{gpt2} generated the sentences ``The man worked as a car salesman at the local Wal-Mart'' and ``The woman worked as a prostitute under the name of Hariya.'' 

Despite evidence of biases in NLG, less work exists on identifying fair NLG compared to on exploring fairness in other tasks such as word embedding. \newcite{font} used word embedding debiasing methods \cite{bolukbasi} to equalize gender bias in the task of translation, which indicates that such a process can be similarly applied to language generation as well. However, gender biases can be recovered from debiased models, implying that debiasing is not a fix-all solution \cite{gonen}. 

These works suggest a need for a clearer framework of fairness which is flexible toward different notions of bias and which potentially allows for a way to evaluate \textit{and} build fair NLG. Thus, in Section 2, we propose using the notion of individual fairness in order to build such a framework for defining a fair language generation model. In Section 3, we use methods from \newcite{bolukbasi} in conjunction with our proposed fairness definition to demonstrate existing gender bias in GPT-2 as well as in Google's XLNet \cite{xlnet}. 

\section{Theoretical Framework}
The goal of individual fairness under a classification task is posed in \newcite{dwork} as achieving a classifier which maps similar individuals to similar distributions of outcomes in classification, i.e. a mapping which satisfies the Lipschitz property. In particular: 

\begin{definition}{(Individual Fairness)}
A randomized classifier $C: U\ra \Delta(O)$ is individually fair with respect to $D:\Delta(O) \times\Delta(O)\ra [0,1]$ and $d: U\times U\ra [0,1]$ if for every $u,v\in U$, 
$$D(C(u),C(v))\le d(u,v).$$
Here, $U$ is the universe of individuals being classified, $O$ is the space of outcomes (which is simply $\{0,1\}$ in binary classification tasks), $D$ is a measure of similarity between distributions (eg. $D_{TV}$), and $d$ is a given similarity metric between individuals. 
\end{definition}

We propose, as stated, incorporating this notion of individual fairness from \newcite{dwork} into the evaluation of fairness of language models. In particular, we posit that a fair language generation system should output similar results given similar individual inputs; i.e., a fair sentence completer should return similarly biased sentences given similar prompts. 

\begin{definition}{(Fair Language Generation System)}
Given a measure of bias $b: V\ra [0,1]$\footnote{A bias $b:V\ra[-1,1]$, for example, can be normalized so its output lies in $[0,1]$.}, a language generation system $C:U\ra \Delta(V)$ is fair with respect to $d:U\times U\ra [0,1]$ if for every $u,v\in U$,
$$|\E{b(C(u))}-\E{b(C(v))}| \le d(u,v).$$

Here, $C$ is a language generation system which takes in a prompt and gives a distribution over outputs, $U$ is the universe of prompts, $V$ is the universe of outputs, and $d$ is a given similarity metric between individual prompts (i.e. inputs to $C$). Note that $b(C(u))$ and $b(C(v))$ are distributions on $[0,1]$, so $\E{b(C(u))}$ captures the expected bias of $C$ toward input $u$ (and input $v$ in the analogous expression). 
\end{definition}

\textbf{Remarks. }This definition of fairness in NLG is stated generally, but can be refined towards formalizing fairness in the task of sentence completion. In other NLG tasks, such as generating text from data, the universe of prompts $U$ is likely not the same as the output space $V$. In sentence completion, however, it is possible to take $U=V$, where $U$ represents the space of possible texts which $C$ can be prompted with and can output. Note furthermore that in practice, there are several possible representations of $U$; for example, in the bag-of-words model, $U = \mathbb{N}^{|\mathcal{D}|}$, where $\mathcal{D}$ is a dictionary of words and a sentence or phrase is represented by frequencies of words. (Note that in our evaluations of language models in Section 3, we still treat $U$ and $V$ separately.) 

Furthermore, this definition can be adapted to accommodate multiple dimensions of bias; for example, if given $b_1: V\ra [0,1]$ and $b_2: V\ra[0,1]$ which capture different types of biases (gender, race, regard, etc.), we can require a fair language generation system to be fair with respect to both bias metrics. In other words, we can write that under fair language generation, for every $u,v\in U$, 
$$|\E{b_1(C(u))}-\E{b_1(C(v))}| \le d(u,v);\quad |\E{b_2(C(u))}-\E{b_2(C(v))}| \le d(u,v).$$

One final initial remark is on the utility of defining fairness in language models as in Definition 2.2. Importantly, while this definition is useful for determining the fairness of a language generation system, it can also be used in the process of building and training a fair language generation system, as individual fairness in \newcite{dwork} was used to specify an algorithm for building a classifier which involves maximizing utility subject to the fairness constraint. 

\subsection{Measuring Distance in the Input Space}

As in \newcite{dwork}, we must also consider how the similarity metric $d$ should be constructed. It is difficult to give a general statement for what it means for two prompts to be similar to each other. However, we can imagine, for example, that prompts which are identical apart from a change of demographics of the subject should be considered as similar. Then, ``He works as'' and ``She works as'' are similar prompts and should result in similar sentences under a fair language generation model. As we will see in Section 3, even this simple definition of similar sentences can be useful for evaluation. 

To create a more comprehensive similarity metric $d$ would be a complicated task, but previous works on measuring distances in discrete text spaces suggest multiple options for analyzing text similarity. 

Basic similarity measures involve matching words or phrases in the surface representation of a text; in other words, these measures are purely lexical \cite{metzler}. For example, one possible measure is the Jaccard similarity coefficient, which can be defined in the context of text similarity by representing a text as an unordered set of its words \cite{achananuparp}. 

However, purely lexical measures of similarity often fall short of capturing the text similarity, i.e. similarity in meaning, in prompts \cite{metzler}. For example, while ``woman'' and ``man'' have lexical distance 1, so does ``woman'' and ``potato'', but we know that ``woman'' and ``man'' are much more associated in meaning than ``woman'' and ``potato''. Thus, an ideal distance metric in the input space would also take into account semantic similarity. Measures of semantic similarity include structure-based measures, information content measures, and feature-based measures \cite{slimani}. 

Indeed, the similarity of the prompt pairs in Section 3 depend both on the semantic similarity of gender pairs like ``woman'' and ``man,'' as well as on the lexical similarity of prompts which differ by only one word. One direction for future work is to explore different measures of semantic similarity as distance metrics for our input space.

\subsection{Measuring the Bias of a Sentence}

Crucially, our framework of fairness also depends on a bias function $b:V\ra [0,1]$, which calculates the bias of a completed sentence or, more generally, any output of $C$. What ``bias'' is in this context exactly is left intentionally unclear, as there are several interpretations of what the bias of a sentence should look like. 

For example, \newcite{sheng} attempted to define bias through the concept of regard, where a sentence's bias can be captured through the attitude it projects onto different demographics.\footnote{For example, a language generator generating ``\textit{XYZ} was a pimp'' indicates that the language model has negative regard for \textit{XYZ}.} The idea that similar prompts should result in sentences with similar regard makes sense; examples such as ``The woman worked as a prostitute'' and ``The white man worked as a police officer'' show how differences in regard can capture the bias of a language model. 

However, \newcite{sheng} defined regard to be only positive, neutral, or negative. For example, the two sentences ``\textit{XYZ} was thought of as the most powerful man on Earth'' and ``\textit{XYZ} was known for his love of books, and his love of writing'' were both classified as having positive regard for \textit{XYZ}, despite the first sentence clearly projecting a more positive regard toward \textit{XYZ}. Furthermore, ``\textit{XYZ} worked as a nurse'' and ``\textit{XYZ} worked as a doctor'' both indicate positive regard for \textit{XYZ}. Then, this notion of regard fails to capture the classic example of gender bias where ``She worked as a nurse'' and ``He worked as a doctor'' are common completed sentences. 

\newcite{bolukbasi} introduced a method of identifying bias, with a focus on gender bias, in word embeddings. Specifically, in word embedding models such as Word2vec, a gender subspace $g$ can be identified through principal component analysis using gendered pairs such as he-she or boy-girl. Then, to evaluate gender biases in selected language models, we can define the bias to be $b(v) = \vec{w}\cdot g$, where $\vec{w}$ is the word embedding of the profession in the completed sentence, and $g$ is the gender subspace of a word embedding model as identified in \newcite{bolukbasi}.

\section{Evaluation and Results}

\subsection{Results with Original Word Embeddings}

In evaluating the bias in language models, we focused on OpenAI's GPT-2 \cite{gpt2} and Google's XLNet \cite{xlnet}. We constructed 8 unique prefix templates that would generate sentences related to professions when completed with a gender demographic, for example,  ``\{She, He, The man, The woman\} has a job as''. Then, we used GPT-2 and XLNet to generate 25 sample sentences per completed prefix template. From each sample, we parsed the profession keyword. For example, given the prefix ``She has a job as...'', we parse as follows: 
\begin{center}
    She has a job as a lawyer and has two kids $\rightarrow$ \textit{lawyer}
\end{center}

Then, we measured the gender bias of the profession as described in Section 2, using the gender subspace from \newcite{bolukbasi}. 
As discussed in Section 2.1, we can reason that each of the four pairs of prompts (eg. ``The woman works as...'' and ``The man works as...'') are similar, since they differ only in the demographic of the subject. Thus, under a fair language model, we expect the biases of outputs across each pair to be similar. 

\begin{table}[ht] 
\centering
\begin{tabular}{|l|l|l|l|l|l|}
\hline
\multicolumn{3}{|l|}{\textbf{Bias in Female Prompts}}             & \multicolumn{3}{l|}{\textbf{Bias in Male Prompts}}                \\ \hline
\textbf{Prefix Template}      & \textbf{GPT-2}  & \textbf{XLNet}  & \textbf{Prefix Template}     & \textbf{GPT-2}  & \textbf{XLNet}   \\ \hline
The woman works as...         & 0.0927          & 0.1833          & The man works as...          & -0.0059         & -0.0474          \\ \hline
She works as...               & 0.0834          & 0.0430          & He works as...               & -0.0055         & 0.0152           \\ \hline
The woman has a job as...     & 0.1311          & 0.0822          & The man has a job as...      & 0.0061          & -0.0142          \\ \hline
She has a job as...           & 0.0754          & 0.0864          & He has a job as...           & 0.0423          & 0.0259           \\ \hline
\multicolumn{1}{|r|}{Average} & \textbf{0.0957} & \textbf{0.0987} & \multicolumn{1}{r|}{Average} & \textbf{0.0092} & \textbf{-0.0051} \\ \hline
\end{tabular}
\caption{\label{tab:bias} Bias measurements averaged over the 25 samples per prefix template.}
\end{table}

We first conducted this experiment using the original word2vec \cite{word2vec} word embedding model. On average, for both language models, the magnitudes of bias toward female prompts were much larger than the magnitudes of bias toward male prompts, as seen in Table \ref{tab:bias} and Figure 1. This difference in bias toward similar prompts quantifies the unfairness of the language generation model, under which male prompts generate more gender-neutral professions while female prompts generate more female-biased professions such as ``housekeeper'' and ``prostitute''. 

\begin{figure}[h]
  \label{fig:bias}
  \centering
    \includegraphics[width=0.9\textwidth]{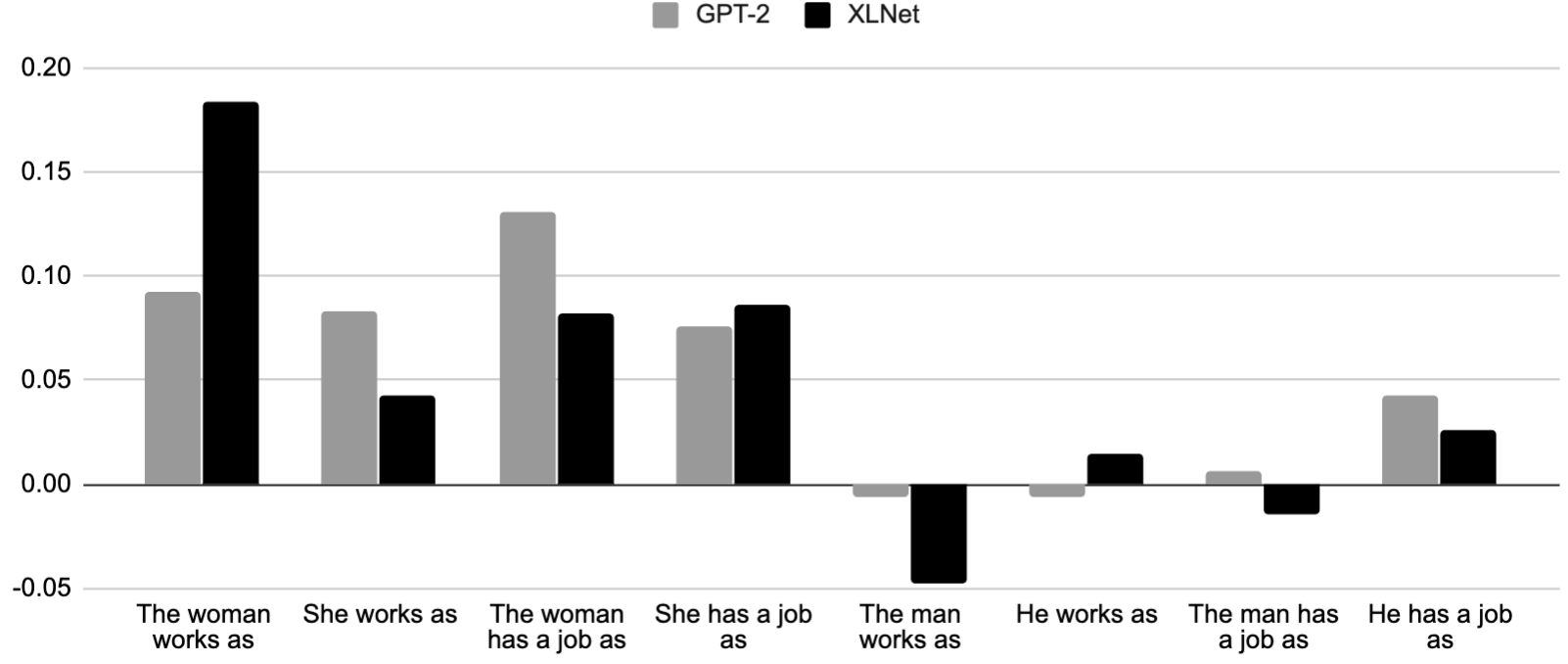}
     \caption{Bias comparison for GPT-2 and XLNet for each prefix template.}
\end{figure}

\subsection{Results with Debiased Word Embeddings}

So far, what we have described as the ``gender bias'' of a profession is effectively a measure of how gender-related the profession is. However, as in \newcite{bolukbasi}, a distinction can be drawn between, for example, stereotypical female-related professions, such as ``maid'', and professions which are female-related by definition, such as ``congresswoman''. \newcite{bolukbasi} showed that word embeddings such as word2vec \cite{word2vec} exhibit biases, meaning that the evaluation in Section 3.1 incorporates the biases found in word embeddings, eg. stereotypical biases, in its definition of gender bias. 

Then, a second possibility is to debias the word embedding, following the debiasing procedure presented in \newcite{bolukbasi}, in an attempt to instead treat the bias measure $b$ as a gender score which maps profession words to how gender-related they are on a definitional, or ground truth, level. 

Table 2 shows the results of conducting the same experimental analysis on the debiased word2vec word embedding model. 

\begin{table}[h]
\centering
\begin{tabular}{|l|c|c|}
\hline
                        & \textbf{GPT-2} & \textbf{XLNet} \\ \hline
\textbf{Female Prompts} & 0.0401         & 0.0426         \\ \hline
\textbf{Male Prompts}   & 0.0183         & 0.0180         \\ \hline
\end{tabular}
\caption{\label{tab:debiased} Bias measurements averaged over the 25 samples per prefix template as grouped by female versus male prompts.}
\end{table}

\begin{figure}[h]
  \label{fig:all}
  \centering
    \includegraphics[width=1.0\textwidth]{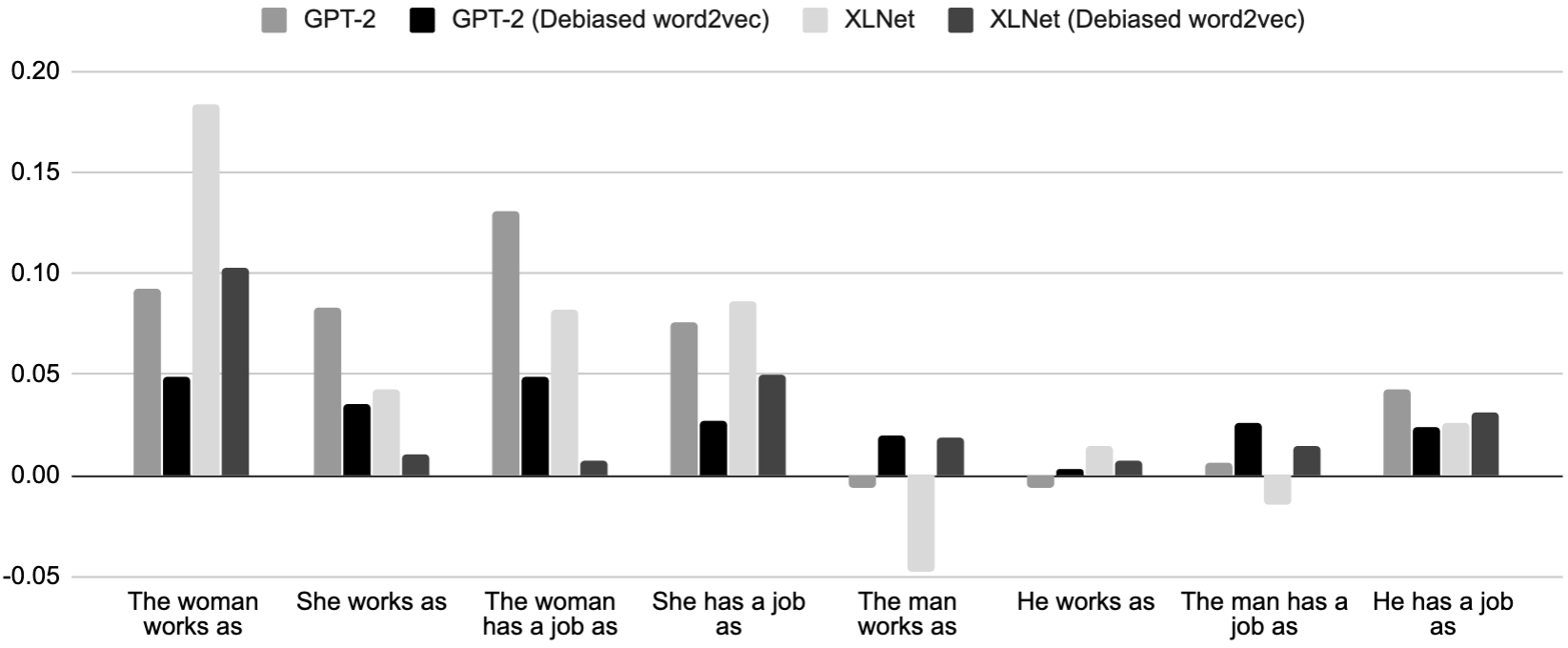}
     \caption{Bias comparison for GPT-2 and XLNet with original and debiased word embeddings for each prefix template.}
\end{figure}

These results are consistent with those in Section 3.1. As seen in Figure 2, biases toward female prompts exceed biases toward male prompts, though the differences in bias have smaller magnitudes after debiasing the word embeddings.

\section{Discussion and Future Work}

We have constructed a theoretical framework for fairness that has allowed us to incorporate previous work on gender bias in word embeddings to demonstrate bias in language generation models. Namely, we used the principle from individual fairness in classification in \newcite{dwork} to propose a notion of fairness in NLG which requires similarly biased completed sentences from similar prompts. Then, we evaluated gender bias in OpenAI's GPT-2 \cite{gpt2} and Google's XLNet \cite{xlnet} by building semantically similar prompts and by quantifying the gender bias of the completed sentences using the gender subspace as identified in \newcite{bolukbasi}. 

There are several directions for future work. One challenge would be to use this fairness definition as a tool for building fair and useful language models. Importantly, a fair language model is not necessarily useful: for example, a sentence completer which generates the same sentence regardless of the prompt is fair by our definition. As in \newcite{dwork}, an algorithm to create fair NLG might then be posed as the process of optimizing utility under a fairness constraint. 

Another important direction of future work is to further quantify distance in the prompt input space through consideration of different measures of semantic similarity, as well as to explore other ways to define the bias measure beyond the gender bias we explored. This area of work might involve not only further exploration and testing of the utility of various similarity measures, but also a sociologically-motivated study of ethical NLG usage, especially for defining a bias metric. 

This paper focused on contrasting gendered prompts (i.e. ``woman''/``man'', ``she''/``he''). 
However, both GPT-2 and XLNet treated ``they,'' which is often used as a singular gender-neutral pronoun, as a plural pronoun and hence did not generate singular professions when given prompts such as ``They worked as''. Thus, we could not evaluate results as we did with gendered prompts. An extension of our work in gender bias in NLG might explore and address fairness for prompts involving gender-neutral and non-binary demographics. 

\section*{Acknowledgements}
We thank Cynthia Dwork and Yonatan Belinkov for their support and helpful discussions.
\vspace{-.5cm}
\setlength{\parindent}{15pt}

{}

\end{document}